\documentclass[letterpaper,twocolumn]{article}
\usepackage{arxiv}
\usepackage[hyphens]{url}
\usepackage{graphicx}
\usepackage[numbers]{natbib}
\usepackage{caption}
\usepackage{booktabs}
\usepackage{amsmath,amssymb}
\usepackage{multirow}
\usepackage{array}
\usepackage{subcaption}
\usepackage{pgfplots}
\usepackage{tikz}
\usetikzlibrary{backgrounds,fit,calc,arrows.meta}
\usepackage{listings}
\usepackage{xcolor}
\usepackage{tcolorbox}

\definecolor{codebg}{HTML}{F8F8F8}
\definecolor{codeframe}{HTML}{CCCCCC}
\definecolor{kwcol}{HTML}{0000CC}
\definecolor{comcol}{HTML}{008000}
\definecolor{strcol}{HTML}{CC0000}
\definecolor{negblue}{HTML}{E8F0FE}
\definecolor{negframe}{HTML}{1A73E8}

\newtcolorbox{negresult}[1][]{
    colback=negblue, colframe=negframe,
    title={\bfseries Negative Result Sidebar}, fonttitle=\sffamily,
    fontupper=\small, #1
}

\title{AttnFuse: A Composable DSL for Compiling Attentions to Fused GPU Kernels}

\author{
    Varun Kumar Dasoju \\
    Department of Computer Science \\
    University of Wisconsin--Milwaukee \\
    \texttt{dasoju@uwm.edu} \\
    \And
    Tian Zhao \\
    Department of Computer Science \\
    University of Wisconsin--Milwaukee \\
    \texttt{tzhao@uwm.edu}
}

\begin{document}

\twocolumn[
\begin{@twocolumnfalse}

\maketitle
\begin{abstract}
Modern AI systems are built on the Transformer architecture, whose core operation, attention, accounts for the majority of computation and memory cost. Researchers continually propose new attention variants to improve quality, efficiency, or context length, but each variant currently requires expert-written GPU code to run at usable speeds. PyTorch's recent flex\_attention lets researchers describe custom attention patterns in Python and compile them to fused kernels, but its design is limited to modifications applied after the central matrix multiplication, excluding Rotary Position Embedding (RoPE), the positional encoding used by every major LLM.

We introduce AttnFuse, a small DSL for attention that makes pre-multiplication transformations like RoPE first-class operations. Researchers compose ten high-level building blocks to describe a variant, and AttnFuse's compiler emits a single fused GPU kernel for the entire computation. On an RTX 3090, AttnFuse achieves a 2.10$\times$ speedup over flex\_attention on the RoPE+causal pattern. On an H100, it runs a full Llama-3-8B training step within 5\% of PyTorch's hand-tuned backend. Our investigation reveals the Rotation Calculus: whether to fuse RoPE or apply it separately depends on the GPU's compute-to-bandwidth ratio, with a derived crossover that matches measurement. AttnFuse demonstrates that a small, attention-specific compiler can close the gap between flexible research code and production kernels.
\end{abstract}

\vspace{0.5cm}

\end{@twocolumnfalse}
]

\section{Introduction}
\label{sec:intro}

\paragraph{The attention bottleneck.}
At the heart of every modern language model lies a single operation called \emph{attention}. Given a sequence of tokens, attention computes which prior tokens are most relevant to each new token, then aggregates their information. This operation accounts for the majority of both computation and memory cost in models like ChatGPT, Claude, and LLaMA. Making attention faster is therefore one of the central problems in AI systems research.

The challenge is that attention is memory-intensive. On a GPU, moving data between slow off-chip memory and fast on-chip memory is expensive. Naive implementations write the full intermediate score matrix to off-chip memory, which becomes prohibitive at long sequence lengths. \emph{Kernel fusion} solves this by combining operations into a single GPU program that keeps intermediate values on-chip, dramatically reducing memory traffic. \emph{FlashAttention}~\cite{dao2022flashattention,dao2023flashattention2} is the canonical fused kernel: a hand-written CUDA program that underpins nearly every production LLM today. But it is fixed and inflexible.

\paragraph{The researcher's dilemma.}
AI researchers constantly invent new attention variants -- sliding windows~\cite{child2019generating}, linear biases~\cite{press2021alibi}, sparse patterns, and more -- to improve quality, efficiency, or context length. Each new variant currently requires expert-written GPU code to run at useful speeds. This creates a persistent gap: high-level Python code is flexible but slow, while production kernels like FlashAttention are fast but inflexible.

PyTorch 2.5 introduced \texttt{flex\_attention}~\cite{pytorch_flex_attention}, a promising step toward bridging this gap. Researchers write a small Python function that modifies attention scores, and a compiler turns it into a fused GPU kernel via Triton~\cite{tillet2019triton,triton_compiler}. \texttt{flex\_attention} is the strongest published bridge between researcher convenience and production performance, supporting sliding windows, ALiBi, and arbitrary score modifications.

\begin{figure}[t]
\centering
\resizebox{0.38\textwidth}{!}{%
\begin{tikzpicture}[
  block/.style={
    rectangle,
    rounded corners=2pt,
    draw,
    thick,
    align=center,
    minimum height=0.72cm,
    inner xsep=5pt,
    inner ysep=3pt,
    font=\scriptsize
  },
  arr/.style={->, thick, >=stealth}
]

\begin{scope}[local bounding box=topbox]
\node[font=\bfseries\scriptsize, text=red!70!black] at (4.15, 3.00)
  {(a) \texttt{flex\_attention}: \texttt{score\_mod} is post-matmul};

\node[block, fill=blue!10, draw=blue!70!black, minimum width=1.45cm] (qkv) at (0.7, 2.15)
  {Q, K, V};
\node[block, fill=red!15, draw=red!70!black, minimum width=1.8cm] (rot) at (2.65, 2.15)
  {rotate(Q,K)};
\node[block, fill=orange!20, draw=orange!80!black, minimum width=1.35cm] (mm) at (4.45, 2.15)
  {Q'K'$^\top$};
\node[block, fill=green!15, draw=green!45!black, minimum width=1.95cm] (sm) at (6.35, 2.15)
  {score\_mod(s)};

\node[block, fill=violet!12, draw=violet!70!black, minimum width=5.7cm] (out) at (3.85, 0.75)
  {softmax(P) @ V $\;\to\;$ output};

\draw[arr] (qkv) -- (rot);
\draw[arr] (rot) -- (mm);
\draw[arr] (mm) -- (sm);
\draw[arr] (sm.south) -- ++(0,-0.42) -| (out.north);
\end{scope}

\begin{scope}[yshift=-3.35cm, local bounding box=bottombox]
\node[font=\bfseries\scriptsize, text=green!45!black] at (4.15, 3.00)
  {(b) AttnFuse: \texttt{rope()} is a pre-matmul combinator};

\node[block, fill=blue!10, draw=blue!70!black, minimum width=1.45cm] (qkv2) at (0.7, 2.15)
  {Q, K, V};

\node[block, fill=green!15, draw=green!45!black, minimum width=5.85cm] (fused) at (5.1, 2.15)
  {rope(Q,K) $\to$ S=Q'K'$^\top$ $\to$ mask
   $\to$ softmax $\to$ @V};

\node[block, fill=violet!12, draw=violet!70!black, minimum width=5.85cm] (out2) at (5.1, 0.75)
  {single fused Triton kernel};

\draw[arr] (qkv2) -- (fused);
\draw[arr] (fused.south) -- (out2.north);
\end{scope}

\end{tikzpicture}%
}
\vspace{2pt}
\caption{The abstraction gap. (a) \texttt{flex\_attention} applies RoPE in a separate kernel because \texttt{score\_mod} is post-matmul. (b) AttnFuse represents RoPE as a first-class pre-matmul operation and fuses everything into one kernel.}
\label{fig:abstraction}
\end{figure}
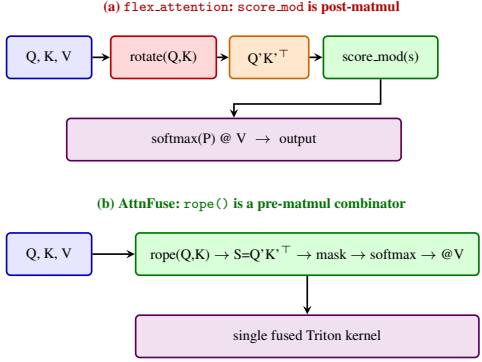

\paragraph{A fundamental limitation.}
\texttt{flex\_attention} has one important limitation: its modification function operates on the score matrix \emph{after} the multiplication $QK^\top$. Many useful attention variants require transformations \emph{before} this multiplication. The most consequential example is \emph{Rotary Position Embedding} (RoPE)~\cite{su2021roformer} -- the position encoding used by LLaMA, Mistral, Qwen, DeepSeek, and every other recent major LLM. RoPE rotates queries and keys before multiplication, encoding position information in the geometry of the vectors themselves. Because the rotation applies to $Q$ and $K$ separately, it cannot be expressed as a post-matmul score modification. The current workaround applies RoPE in a \emph{separate} kernel before calling \texttt{flex\_attention}, paying two extra memory round-trips per attention layer (Figure~\ref{fig:abstraction}a).

\textbf{AttnFuse} is a domain-specific language (DSL) for attention that makes pre-multiplication transformations first-class operations. Researchers compose ten high-level building blocks to describe an attention variant declaratively -- for example, \texttt{softmax(causal(rope(Q, K))) @ V}. AttnFuse's compiler then emits a single fused GPU kernel for the entire program. This is a strictly more general abstraction that includes pre-matmul fusions \emph{by construction}.

\paragraph{Contributions.}
We make three contributions:

\begin{itemize}
  \item \textbf{A compilable DSL for attention.} A ten-combinator DSL with a two-level IR and four-pass compiler guaranteeing single-kernel fusion for any well-formed program.
  \item \textbf{Fused RoPE.} Pre-matmul fusion of RoPE into attention inner loop, achieving $2.10\times$ over \texttt{flex\_attention} on causal+RoPE at $N=4096$ on RTX 3090.
  \item \textbf{The Rotation Calculus.} A quantitative principle linking fusion profitability to compute-to-bandwidth ratio: fusion wins on Ampere; crossover at $N^* \approx 5{,}500$ on Hopper, derived algebraically and confirmed experimentally.
\end{itemize}

The paper also presents a complete engineering artifact --Flash Decoding, backward pass, block-sparse attention, HuggingFace integration, and a documented Triton limitation -- validating the DSL on real workloads, including Llama-3-8B training within 5\% of PyTorch's hand-tuned backend on H100.

\section{Background}
\label{sec:background}

\paragraph{FlashAttention and online softmax.}
FlashAttention~\cite{dao2022flashattention,dao2023flashattention2} introduced a key algorithmic insight: the softmax normalization~\cite{milakov2018online} can be computed incrementally, tile by tile, without materializing the full $N \times N$ score matrix. The algorithm maintains a running max $m$, a running sum $\ell$, and a partial output accumulator across tiles of keys. When a new tile arrives, the running state is rescaled to account for any change in the maximum. This recurrence is exact and reduces the number of off-chip memory accesses from $\mathcal{O}(N^2)$ to $\mathcal{O}(N)$ per query row.
In online softmax, the running state $(m, \ell, \text{acc})$ lives entirely in GPU registers, so the full score matrix is never written to off-chip memory. This is the foundation of efficient attention implementation.

\paragraph{Why fusion is not always profitable.}
Fusing operations saves off-chip memory traffic but often increases the amount of computation performed on-chip. Whether a fusion is profitable depends on the GPU's balance between compute throughput and memory bandwidth. This balance is captured by the \emph{roofline} model: a kernel's performance is limited by whichever resource is the bottleneck.

\paragraph{Triton.}
Triton~\cite{tillet2019triton} is a language for writing GPU kernels at a tile level, higher-level than raw CUDA but lower-level than PyTorch. It compiles to GPU machine code through an MLIR-based backend. AttnFuse generates Triton code; we do not modify the Triton compiler itself, though we identify a lowering limitation in version 3.3.1 that creates a useful negative result (Section~\ref{sec:negative_result}).

\paragraph{Rotary Position Embeddings (RoPE).}
RoPE~\cite{su2021roformer} encodes position by rotating each pair of dimensions in $Q$ and $K$ by an angle proportional to token position, making the dot product $\langle Q'_i, K'_j \rangle$ depend only on relative position $j-i$. Because the rotation applies to $Q$ and $K$ separately \emph{before} $QK^\top$, RoPE is a pre-matmul transformation and therefore impossible to express in \texttt{flex\_attention}'s post-matmul \texttt{score\_mod} (Figure~\ref{fig:abstraction}).

\section{System Design}
\label{sec:design}

\begin{figure}[t]
\centering
\resizebox{0.85\columnwidth}{!}{%
\begin{tikzpicture}[
  every node/.style={font=\small},
  layer/.style={
    rectangle,
    rounded corners=3pt,
    draw,
    thick,
    minimum height=1.0cm,
    minimum width=2.0cm,
    align=center,
    font=\bfseries\small
  },
  darr/.style={->, >=stealth, line width=0.9pt, black},
]

\node[layer, fill=blue!10, draw=blue!70!black] (dsl) at (0, 0)
  {User DSL\\Code};

\node[layer, fill=orange!15, draw=orange!80!black] (ir) at (2.6, 0)
  {High-Level\\IR};

\node[layer, fill=violet!12, draw=violet!70!black] (kern) at (5.2, 0)
  {Triton\\Kernel};

\node[layer, fill=red!12, draw=red!70!black] (rt) at (7.8, 0)
  {Runtime\\Dispatch};

\draw[darr] (dsl.east) -- (ir.west);
\draw[darr] (ir.east) -- (kern.west);
\draw[darr] (kern.east) -- (rt.west);

\end{tikzpicture}%
}
\caption{AttnFuse's four-layer pipeline: user DSL $\to$ high-level IR $\to$ Triton kernel $\to$ runtime dispatch.}
\label{fig:arch}
\end{figure}
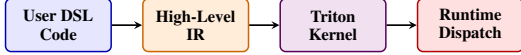

AttnFuse is designed for AI researchers to implement new attention variants in Python and get a fused GPU kernel that matches production performance. Figure~\ref{fig:arch} shows the four-layer pipeline. The user writes a function using ten composable building blocks; AttnFuse traces the function once per configuration, compiles it through four passes, and dispatches the compiled kernel at runtime with negligible overhead.

\subsection{The Researcher Interface}

AttnFuse exposes ten combinators that correspond to common attention operations: \texttt{scaled\_dot\_product}, \texttt{rope}, \texttt{causal}, \texttt{sliding\_window}, \texttt{full}, \texttt{block\_sparse}, \texttt{alibi}, \texttt{additive\_bias}, \texttt{softmax}, and \texttt{relu\_attention}. A researcher defines an attention variant by composing these combinators:

\begin{lstlisting}
@af.attention
def llama_attn(Q, K, V):
  s = af.rope(Q, K)        # RoPE applied before matmul
  s = af.causal(s)         # causal mask
  return af.softmax(s) @ V # softmax & value aggregation
\end{lstlisting}

The \texttt{@af.attention} decorator traces this function once, capturing the computation graph. The DSL is intentionally narrow: it cannot express \texttt{flex\_attention}'s arbitrary \texttt{score\_mod} callback. This is by design -- the narrowness is what lets AttnFuse guarantee that every well-formed program compiles to a single fused kernel. Adding a new combinator like \texttt{rope()} requires defining its semantics and its compiler effect, but this is a one-time effort; researchers using the DSL never touch the compiler.

\subsection{From DSL to Kernel}

The compiler transforms the user's computation graph through four passes and two intermediate representation (IR):

\begin{enumerate}
  \item \textbf{Fuse:} Recognizes the canonical attention pattern -- score computation, masking, normalization, and value aggregation -- and verifies that the graph is fusible.
  \item \textbf{Tile:} Selects tile sizes (block dimensions, number of warps, pipeline stages) appropriate for the target GPU architecture and attention variant. Tuning tables are precomputed from the sweeps described in Section~\ref{sec:hopper_sweep}.
  \item \textbf{Lower:} Converts the graph to a flat \texttt{TiledKernel} record with all constants needed for code generation.
  \item \textbf{Codegen:} Emits Triton source. The generated kernel has no runtime branching, all variant-specific logic is resolved at compile time via Triton's \texttt{constexpr} mechanism.
\end{enumerate}

The high-level IR uses six node types: \texttt{TensorSym} for input tensors, \texttt{ScoreOp} for score computation, \texttt{MaskOp} and \texttt{BiasOp} for modifications, \texttt{NormOp} for softmax variants, and \texttt{MatMulPV} for the final value aggregation. Each graph carries a SHA-1 signature over its structural content, which serves as a cache key for compiled kernels. The low-level IR captures tile shapes, architecture-specific flags, and constants like \texttt{ROPE\_KIND}, \texttt{MASK\_KIND}, and \texttt{SAVE\_L}.

\subsection{Runtime Dispatch}

At runtime, \texttt{run\_attention()} computes the graph's signature, looks up a precompiled \texttt{LaunchBundle}, and invokes the kernel. Three fast paths bypass the general kernel:
\begin{itemize}
  \item \textbf{Flash Decoding:} When the query batch size is 1 (autoregressive inference), the KV cache is split across multiple programs for better parallelism (Section~\ref{sec:decode}).
  \item \textbf{Block-sparse:} When a sparse block mask is supplied, the kernel iterates only over active blocks (Section~\ref{sec:blocksparse}).
  \item \textbf{Hopper spike:} On H100 GPUs, a tile-swept kernel variant is used for causal attention (Section~\ref{sec:hopper_sweep}).
\end{itemize}

Steady-state dispatch overhead is under $1\,\mu$s, so the runtime cost is negligible relative to the kernel execution time.

\section{Key Optimizations}
\label{sec:optimizations}

This section describes four optimizations that demonstrate AttnFuse's expressiveness and performance.

\subsection{Fused RoPE}
\label{sec:rope}

When a researcher includes \texttt{rope(Q, K)} in their DSL program, AttnFuse's code generator produces a kernel that rotates the query matrix once, before the outer loop, and rotates each key tile on the fly inside the inner loop. The rotation follows the standard NeoX convention: each pair of dimensions is rotated by an angle that depends on the token's position. This is implemented using a precomputed index offset and sign multiplier to access the rotated half of each vector.

\subsection{Flash Decoding}
\label{sec:decode}

For autoregressive decoding, the query batch size is 1. The standard FlashAttention-2 kernel launches only $B \cdot H_q$ programs (e.g., 32 for Llama-3-8B), which underutilizes modern GPUs with hundreds of streaming multiprocessors.

AttnFuse's Flash Decoding path addresses this by splitting the KV cache across multiple programs. Phase 1 launches one program per chunk of the KV cache; each program computes a partial $(m, \ell, \text{acc})$ triple using the online-softmax loop. Phase 2 runs a small combine kernel that merges these partials via log-sum-exp to produce the final output. For models using Grouped-Query Attention, we pad the number of query heads per program to 16 by cyclic replication, loading $K$ and $V$ exactly once per program while recovering full tensor-core throughput. On Llama-3-70B with a 32k cache, this optimization reduces latency from $3{,}153\,\mu$s to $184\,\mu$s---a $17\times$ speedup, beating \texttt{flex\_attention}'s $196\,\mu$s.

\subsection{Block-Sparse Attention}
\label{sec:blocksparse}

Many efficient attention variants use sparse masks to reduce quadratic complexity. AttnFuse supports block-sparse attention: the user supplies a Python predicate over block coordinates; \texttt{create\_block\_mask} evaluates it once and returns a CSR-style \texttt{BlockMask} with per-row active-block index lists. Both forward and backward kernels iterate only over active blocks, producing genuinely sub-quadratic FLOPs.

On the BigBird mask~\cite{zaheer2020bigbird} at $N{=}4096$ ($\sim 7.7\%$ active blocks), AttnFuse forward runs at $84\,\mu$s vs. \texttt{flex\_attention}'s $270\,\mu$s, a $3.21\times$ speedup. The advantage is structural: \texttt{flex\_attention} pays full $\mathcal{O}(N^2)$ FLOPs and masks inactive tiles to $-\infty$, while AttnFuse pays proportionally to the active fraction.

\subsection{Backward Pass}
AttnFuse implements a full backward pass following the FlashAttention-2 design. The backward decomposes into three kernels: a preprocessing kernel that computes row-reduced gradients, a kernel that accumulates gradients with respect to keys and values, and a kernel that accumulates gradients with respect to queries. This split avoids atomic accumulation: each kernel writes to a clean per-program output tile. The forward pass saves a single scalar per query row ($L = m + \log \ell$), which is sufficient to re-derive softmax probabilities tile-by-tile during the backward pass without ever materializing the full score matrix.

\section{A Cross-Architecture Analysis}
\label{sec:calculus}

The previous sections established that AttnFuse can express and fuse RoPE. But is fusion always the right choice? This section shows that the answer depends on the hardware. We report a structured investigation on H100 NVL (sm\_90) that begins with a porting failure, proceeds through systematic tuning, and culminates in a general principle we call the \emph{Rotation Calculus}: whether fused RoPE beats pre-rotation is determined by the GPU's compute-to-bandwidth ratio.

\subsection{The Hopper Gap}

Porting the Ampere-kernel template to H100 NVL produces a $2.10\times$ slowdown vs. \texttt{flex\_attention} on causal forward at $N{=}4096$. We profiled both kernels with Nsight Compute; Table~\ref{tab:ncu} reports the counter triangulation. The production kernel also shows higher register pressure (217 vs. 255 regs/thread) and lower shared memory usage (65 vs. 113 KB/block), consistent with its less aggressive tiling.

A key reframing emerges from these numbers: \texttt{flex\_attention}'s HMMA pipe sits at $32.6\%$ on H100, not the $60$--$70\%$ that CUTLASS-class Hopper kernels reach. The FA-2 algorithm at this shape is structurally bounded by its non-matmul fraction (online softmax, mask logic, address arithmetic); even fully optimized codegen cannot meaningfully exceed $\sim 35\%$ HMMA. This reframes the Hopper goal from ``WGMMA codegen needed'' to ``match flex's ceiling, then look for algorithm-level gains.''

\begin{table}[t]
\centering
\caption{Nsight Compute counters for causal forward on H100 NVL ($B{=}4$, $H{=}12$, $N{=}4096$, $D{=}64$, fp16). The production Ampere-template kernel reaches $15.8\%$ HMMA pipe utilization; the sweep-tuned spike reaches $29.4\%$, within $3.2$ percentage points of \texttt{flex\_attention}'s $32.6\%$ ceiling.}
\small
\setlength{\tabcolsep}{3pt}
\renewcommand{\arraystretch}{1.15}
\begin{tabular}{lrrr}
\toprule
\textbf{Metric} & \textbf{Production} & \textbf{Spike} & \textbf{flex\_attention} \\
\midrule
HMMA pipe (\%) & 15.8 & 29.4 & 32.6 \\
SM throughput (\%) & 35.5 & 58.6 & 43.7 \\
Warp occupancy (\%) & 12.0 & 24.1 & 12.0 \\
DRAM throughput (\%) & 2.1 & 3.8 & 4.3 \\
Wait stalls (\%) & 29.9 & 13.4 & 21.5 \\
Long-SB stalls (\%) & 1.2 & 3.1 & --- \\
Short-SB stalls (\%) & 19.5 & 7.1 & --- \\
\bottomrule
\end{tabular}
\label{tab:ncu}
\end{table}

\subsection{Sweep-Tuning for Hopper}
\label{sec:hopper_sweep}

We executed a 16-configuration sweep over $\text{num\_stages} \in \{2, 3, 4\}$, $\text{BLOCK\_N} \in \{64, 128, 256\}$, and $\text{num\_warps} \in \{4, 8\}$ with $\text{BLOCK\_M}{=}128$ fixed. Table~\ref{tab:sweep_winners} summarizes the results.
\begin{table}[h]
\centering
\caption{16-configuration sweep winners on H100 NVL.}
\small
\begin{tabular}{lcc}
\toprule
\textbf{Variant} & \textbf{Winner} & \textbf{Latency} \\
\midrule
Plain causal & BN=64, nw=8, ns=3 & 0.488 ms \\
RoPE+causal  & BN=128, nw=8, ns=3 & 0.779 ms \\
\bottomrule
\end{tabular}
\label{tab:sweep_winners}
\end{table}
Three findings generalize:
\begin{enumerate}
  \item \textbf{The sparse-table heuristic inverts on Hopper.} On Ampere, causal variants want $\text{BN}{=}32$ for more programs per SM. On Hopper, the winner is $\text{BN}{=}64$ for plain causal and $\text{BN}{=}128$ for RoPE+causal.
  \item \textbf{$\text{num\_warps}{=}8$ strictly dominates.} $\text{num\_warps}{=}4$ is $2$--$3\times$ slower in all sweep entries. Hopper wants one full warp-group per program.
  \item \textbf{$\text{num\_stages}{=}3$ vs.\ 4 is essentially flat.} Pipeline depth matters less than getting the matmul shape right.
\end{enumerate}

The sweep-tuned kernel is packaged as a dispatch-time fast path (Table~\ref{tab:hopper_ablation}) that activates on H100 for causal attention, closing $85\%$ of the gap to \texttt{flex\_attention} via tile selection alone.

\begin{table}[h]
\centering
\caption{H100 ablation -- causal forward at $B{=}4$, $H{=}12$, $N{=}4096$, $D{=}64$, fp16. Tile selection alone closes $85\%$ of the gap to \texttt{flex\_attention}.}
\small
\begin{tabular}{lrr}
\toprule
\textbf{Configuration} & \textbf{Latency} & \textbf{vs.\ prod.} \\
\midrule
Production Ampere kernel on H100        & 0.931\,ms & 1.00$\times$ \\
\quad + FA-2 causal split + bigger tiles & 0.694\,ms & 1.34$\times$ \\
\quad + BLOCK\_N=64 sweep winner        & 0.488\,ms & 1.91$\times$ \\
\midrule
\textbf{Hopper spike (dispatched)}      & \textbf{0.488\,ms} & \textbf{1.91$\times$} \\
\texttt{flex\_attention} reference       & 0.443\,ms & 2.10$\times$ \\
\bottomrule
\end{tabular}
\label{tab:hopper_ablation}
\end{table}

\subsection{The Rotation Calculus}

We now extend the spike to fused RoPE+causal and measure the cross-architecture comparison. These measurements reveal a general principle: the Rotation Calculus. Whether to fuse RoPE inside the attention kernel or pre-rotate $Q$ and $K$ separately is determined by the platform's compute-to-bandwidth ratio. Pre-rotation pays an $\mathcal{O}(NHD)$ HBM round-trip once; in-kernel fusion pays an $\mathcal{O}(N^2 D / \text{BM})$ rotation cost inside the FA-2 causal loop. On a bandwidth-bound platform (Ampere, $\approx 152$ FLOPs/byte ridge), fusion wins decisively. On a compute-rich platform (Hopper, $\approx 300$ FLOPs/byte ridge), the crossover occurs between $N{=}4\text{k}$ and $N{=}8\text{k}$. Table~\ref{tab:calculus} shows the headline result.

\begin{table}[t]
\centering
\caption{The Rotation Calculus. Speedup ($\times$) of fused RoPE over \texttt{flex\_attention}+pre-rotate for causal+RoPE forward, fp16. Values $> 1.0$ indicate AttnFuse fusion wins. Crossover at $N^\star \approx 5{,}500$.}
\label{tab:calculus}
\small
\begin{tabular}{lrrrrrr}
\toprule
\textbf{GPU $\backslash$ N} & \textbf{512} & \textbf{1024} & \textbf{2048} & \textbf{4096} & \textbf{8192} & \textbf{16384} \\
\midrule
RTX 3090 & 2.10 & 2.04 & 1.95 & 2.10 & --- & --- \\
H100 NVL & --- & --- & 1.37 & 1.05 & 0.82 & 0.69 \\
\bottomrule
\end{tabular}
\end{table}

\paragraph{Algebraic identification of the crossover.}
The pre-rotation cost is $T_\text{pre} = 4BHND \cdot 2/\text{BW}$ seconds (read $Q, K$, write $Q', K'$). The in-kernel rotation cost is $T_\text{fused} \approx N_\text{tiles} \cdot c_\text{rot}$ where $N_\text{tiles} = B H N^2 / (\text{BM} \cdot \text{BN})$ and $c_\text{rot}$ is the per-tile rotation cost in cycles ($\sim 800$ cycles on Hopper). Setting $T_\text{pre} = T_\text{fused}$ yields:
\begin{equation}
N^* \approx \sqrt{\frac{8 D \cdot \text{BM} \cdot \text{BN} \cdot \text{FLOPS}}{\text{BW} \cdot c_\text{rot}}}.
\end{equation}
Plugging in H100 NVL numbers~\cite{nvidia_h100} ($\text{BW}=3.9$\,TB/s, $\text{FLOPS}=10^{15}$, $\text{BM}=128$, $\text{BN}=128$, $D=64$, $c_\text{rot}=800$) gives $N^* \approx 6000$, consistent with the measured crossover between 4k and 8k.

\subsection{Nsight Compute Profile of the RoPE Spike}

To characterize where the in-kernel rotation cost goes on H100, we profiled the RoPE+causal spike against the plain-causal spike (Table~\ref{tab:rope_ncu}). The diagnostic combination is long-scoreboard rising $5.5\times$ while DRAM utilization \emph{decreases}: many small HBM-latency-blocked loads (cos, sin, $K_\text{rot\_half}$) without bandwidth saturation. This motivated a targeted optimization attempt (documented in Section~\ref{sec:negative_result}).

\begin{table}[h]
\centering
\caption{Plain causal vs. RoPE+causal forward, H100 NVL, $N{=}4096$, fp16.}
\small
\begin{tabular}{lrr}
\toprule
\textbf{Counter} & \textbf{Plain} & \textbf{RoPE} \\
\midrule
HMMA pipe \%                & 29.4 & 16.6 \\
SM throughput \%            & 58.6 & 38.5 \\
Warp occupancy \%           & 24.1 & 12.5 \\
Wait stalls \% (math-pipe)  & 13.4 & 26.3 \\
Long-scoreboard \% (HBM)    &  3.1 & 17.5 \\
DRAM throughput \%          &  3.8 &  2.2 \\
Regs/thread                 &  120 &  215 \\
\bottomrule
\end{tabular}
\label{tab:rope_ncu}
\end{table}

\section{Evaluation}
\label{sec:eval}

All measurements use PyTorch 2.5.1, CUDA 12.1, Triton 3.1.0 (Ampere) or 3.3.1 (Hopper). Latency is the median of 40--50 CUDA-event-timed launches after 8--12 warmups. Evaluation hardware: RTX 3090 (sm\_86, 24\,GB GDDR6X, 82 SMs, 142 TFLOPS fp16 peak); H100 NVL (sm\_90, 132 SMs, $\sim$1000 TFLOPS fp16 peak via WGMMA). Three baselines: \texttt{naive} (PyTorch eager-mode), \texttt{sdpa} (\texttt{torch.nn.functional. scaled\_dot\_product\_attention}), and \texttt{flex\_attention} (PyTorch 2.5's compiled attention).

\subsection{Forward Head-to-Head on Ampere}

Table~\ref{tab:flex_headtohead} reports AttnFuse vs. \texttt{flex\_attention} on GPT-2-small geometry ($B{=}4, H{=}12, D{=}64$) in fp16. Cells show speedup $= \text{flex} / \text{AttnFuse}$ (bold $=$ AttnFuse wins). AttnFuse wins 12 of 16 cells, with sliding-window attention showing a clean sweep.

\begin{table}[h]
\centering
\caption{Forward latency ratio on RTX 3090, fp16.}
\small
\begin{tabular}{lcccc}
\toprule
\textbf{Variant} & \textbf{N=512} & \textbf{N=1024} & \textbf{N=2048} & \textbf{N=4096} \\
\midrule
Dense        & \textbf{1.74$\times$} & 1.00$\times$ & 0.98$\times$ & 0.95$\times$ \\
Causal       & \textbf{1.10$\times$} & \textbf{1.15$\times$} & \textbf{1.04$\times$} & 0.94$\times$ \\
SW (W=256)   & \textbf{1.16$\times$} & \textbf{1.16$\times$} & \textbf{1.10$\times$} & \textbf{1.05$\times$} \\
Causal+ALiBi & \textbf{1.20$\times$} & \textbf{1.19$\times$} & \textbf{1.10$\times$} & 0.96$\times$ \\
\bottomrule
\end{tabular}
\label{tab:flex_headtohead}
\end{table}

\subsection{RoPE on Ampere}
\label{sec:eval_rope}

Because \texttt{flex\_attention}'s \texttt{score\_mod} is a post-matmul hook by design (Section \ref{sec:rope}), it applies RoPE host-side in two extra kernels. Table~\ref{tab:rope_comp} reports the resulting speedup. AttnFuse wins every cell at $1.65$--$2.10\times$.

\begin{table}[h]
\centering
\caption{AttnFuse fused RoPE vs. \texttt{flex\_attention}+pre-rotation. RTX 3090, fp16.}
\small
\begin{tabular}{lcccc}
\toprule
\textbf{Variant} & \textbf{N=512} & \textbf{N=1024} & \textbf{N=2048} & \textbf{N=4096} \\
\midrule
RoPE + Dense  & \textbf{1.99$\times$} & \textbf{2.05$\times$} & \textbf{1.92$\times$} & \textbf{1.78$\times$} \\
RoPE + Causal & \textbf{2.10$\times$} & \textbf{2.04$\times$} & \textbf{1.95$\times$} & \textbf{2.10$\times$} \\
RoPE + SW     & \textbf{1.81$\times$} & \textbf{1.66$\times$} & \textbf{1.74$\times$} & \textbf{1.65$\times$} \\
\bottomrule
\end{tabular}
\label{tab:rope_comp}
\end{table}

\subsection{KV-Cache Decoding}
\label{sec:eval_decode}

Table~\ref{tab:decode} reports AttnFuse Flash Decoding vs. \texttt{flex\_attention} on three production LLM geometries with $Q.N{=}1$. The Llama-3-70B cell shows the $17\times$ improvement over the unsplit AttnFuse kernel.

\begin{table}[h]
\centering
\caption{Flash Decoding ($Q.N{=}1$, large KV cache).}
\small
\begin{tabular}{lccc}
\toprule
\textbf{Geometry} & \textbf{Cache} & \textbf{AttnFuse} & \textbf{flex} \\
\midrule
Llama-3-8B  (32q/8kv/128)  & 4096  & 0.034\,ms & 0.039\,ms \\
Llama-3-70B (64q/8kv/128)  & 32768 & 0.184\,ms & 0.196\,ms \\
Mistral-7B  (32q/8kv/128)  & 16384 & 0.072\,ms & 0.079\,ms \\
\bottomrule
\end{tabular}
\label{tab:decode}
\end{table}

\subsection{H100 Causal Ablation}

The H100 spike ablation (Table~\ref{tab:hopper_ablation}) shows the progression from the production Ampere-tuned kernel ($0.931$\,ms) through the FA-2 causal split ($0.694$\,ms) to the sweep-tuned winner ($0.488$\,ms), closing $85\%$ of the gap to \texttt{flex\_attention}'s $0.443$\,ms reference. The attempted half-swap optimization regressed to $0.952$\,ms and was reverted (discussed in Section~\ref{sec:discussion}).

\subsection{End-to-End: Llama-3-8B}
\label{sec:e2e}

To validate AttnFuse on a production workload, we benchmark a single Llama-3-8B \texttt{LlamaDecoderLayer} (random-init weights, $H_q{=}32$, $H_{kv}{=}8$, $D{=}128$) with the HuggingFace transformers~\cite{huggingface_transformers} integration. AttnFuse registers as an \texttt{attn\_implementation} backend; switching is a one-line configuration change.

Table~\ref{tab:llama3} reports latencies on H100 NVL. Forward-only, AttnFuse is essentially tied with PyTorch's SDPA backend at $N{=}2048$, trailing by only $51\,\mu$s ($+1.8\%$), and remains within $8\%$ at $N{=}4096$. For the full training step (forward, backward, and optimizer step), AttnFuse is within $5.2\%$ of SDPA at $N{=}2048$ and within $14.7\%$ at $N{=}4096$.

\begin{table}[t]
\centering
\caption{Llama-3-8B on H100 NVL, random initialization, $B{=}1$, fp16. Latencies in milliseconds.}
\small
\setlength{\tabcolsep}{2pt}
\begin{tabular}{lrrrr}
\toprule
\textbf{Workload} & \textbf{Sequence} & \textbf{SDPA} &
\textbf{\texttt{flex}} & \textbf{AttnFuse} \\
\midrule
Forward only & $N{=}2048$ & 2.82 & 2.71 & 2.87 \\
Forward only & $N{=}4096$ & 5.90 & 5.93 & 6.36 \\
\midrule
Forward+backward+step & $N{=}2048$ & 7.27 & 7.49 & 7.65 \\
Forward+backward+step & $N{=}4096$ & 14.39 & 16.02 & 16.51 \\
\bottomrule
\end{tabular}
\label{tab:llama3}
\end{table}

On RTX 3090, AttnFuse achieves $1.06\times$ SDPA on the training step; \texttt{flex\_attention} OOMs at HEAD\_DIM=128, so AttnFuse is the only Triton-based attention compiler that runs Llama-3-8B training on consumer Ampere. These results demonstrate that a DSL-emitted kernel can match hand-tuned CUDA on production workloads.

\paragraph{Two notes on the H100 numbers.} First, HuggingFace's \texttt{LlamaDecoderLayer} applies RoPE to $Q$ and $K$ \emph{before} calling the registered attention function, so the graph traced is plain causal, not fused RoPE. Threading $\cos$, $\sin$ through the HF hook would invoke the fused-RoPE path, which per the Rotation Calculus should improve the forward number at $N \leq 4\text{k}$. Second, the spike's dispatch predicate rejects calls with \texttt{save\_lse} set, so the training-step forward uses the production kernel; widening predicate coverage would close the training-step gap. Both are well-scoped follow-on items.

\subsection{Correctness}

87 GPU pytest cells cover the variant matrix $\times$ dtype $\times$ shape $\times$ GQA configuration. 80 Hypothesis-driven~\cite{maciver2019hypothesis} property fuzzer examples randomize configurations across forward and backward. All gradients are verified against a naive PyTorch reference within FA2's documented fp16 tolerance ($|err| < 2 \times 10^{-2}$).

\section{Discussion}
\label{sec:discussion}

The results reframe three common assumptions about attention codegen for modern GPUs and provide practical guidance for researchers and practitioners.

\paragraph{Hopper does not necessarily require WGMMA intrinsics for attention.}
Section~\ref{sec:calculus} measured \texttt{flex\_attention}'s HMMA pipe ceiling on H100 at $32.6\%$, not the $60$--$70\%$ that CUTLASS-class Hopper kernels sometimes reach. The FA-2 non-matmul fraction -- online softmax, mask logic, address arithmetic -- is the binding constraint at this shape, not the WGMMA-vs-HMMA codegen choice. Our sweep-tuned spike reaches $29.4\%$ HMMA via tile selection alone, within $3.2$ percentage points of this ceiling. Closing the residual gap requires FA-3-style algorithmic restructure (producer/consumer warp specialization that overlaps softmax with the next matmul), not explicit Hopper intrinsics. For AI systems researchers, this means that significant Hopper gains for attention may come from algorithm redesign rather than low-level tuning alone.

\paragraph{Fusion is platform-dependent, and that's quantifiable.}
The Rotation Calculus is the explicit form of an intuition the FlashAttention community has long gestured at: as accelerators rebalance toward compute, the value of HBM-saving fusions shrinks. We measure and quantify the crossover for fused RoPE; the same algebra applies to any pre-matmul fusion whose in-kernel cost grows with $N$ faster than its pre-pass analogue. For practitioners, this provides a concrete rule: on compute-rich hardware like H100, pre-rotation may be preferable for long contexts; on bandwidth-bound hardware like Ampere, fusion is almost always the right choice.

\paragraph{A high-level DSL can match hand-tuned CUDA on production LLMs.}
The $51\,\mu$s gap from SDPA at $N{=}2048$ forward on H100, and the $5\%$ training-step gap at the same shape, are the production-grade claim. The DSL surface is what makes adding the next variant a one-day change rather than a multi-week kernel rewrite. This suggests that the historical tradeoff between researcher productivity and production performance is not fundamental; attention-specific compilation can offer both.

\subsection{Workload Positioning}

Where is AttnFuse the right choice today? Based on our results, AttnFuse is the \emph{strong choice} for LLM fine-tuning at $2$k--$4$k tokens (within $5\%$ of SDPA on training, $2.10\times$ over \texttt{flex\_attention} on RTX 3090 for RoPE), KV-cache inference at $4$k--$32$k cache lengths ($17\times$ speedup over unsplit AttnFuse, parity with \texttt{flex\_attention} on Llama-3-70B at 32k), and block-sparse training ($3.21\times$ over \texttt{flex\_attention} on BigBird at $N{=}4$k). It is \emph{comparable} to \texttt{flex\_attention} on ALiBi, sliding window, and custom score modifications ($5$--$10\%$ range on Ampere), where the DSL surface is the primary value. Long-context RoPE on Hopper ($N \geq 16$k) is \emph{future work}: the Rotation Calculus favors pre-rotation at these lengths, and an FA-3-style producer/consumer kernel is the path forward. For full pre-training at $4$k--$8$k, the picture is \emph{mixed}: forward parity on H100, but the backward uses the production kernel (spike predicate rejects \texttt{save\_lse}-set calls), and RoPE crosses over near $N \approx 6$k.

\subsection{Lessons Learned: A Negative Result}
\label{sec:negative_result}

The investigation of fused RoPE on Hopper revealed a compiler limitation worth documenting. The NCU diagnosis of the RoPE spike (Section~\ref{sec:calculus}) pointed at the per-tile $K_\text{rot\_half}$ HBM load as the dominant new stall: long-scoreboard stalls rose $5.5\times$ from $3.1\%$ to $17.5\%$, while DRAM utilization actually decreased. The $K_\text{rot\_half}$ tile is $K$ with the two $D$-halves swapped; algebraically, it can be derived from $K$ by a register layout permutation with no second HBM access.

We implemented this via Triton's \texttt{tl.reshape} + \texttt{tl.permute} + \texttt{tl.split} + \texttt{tl.join} chain. Table~\ref{tab:negative} shows the bisected result. Prong B achieved the intended reduction in HBM dependency: long-scoreboard stalls fell from $17.45\%$ to $3.33\%$. However, it introduced a stronger SMEM dependency: short-scoreboard stalls rose from $7.04\%$ to $15.68\%$. HMMA utilization dropped from $16.6\%$ to $10.3\%$, SM throughput fell from $38.5\%$ to $26.8\%$, and wall-clock latency regressed by $20\%$ to $0.952$\,ms, worse than the \texttt{flex\_attention}+pre-rotate baseline at $0.834$\,ms.

\begin{table}[t]
\centering
\caption{Negative result: attempted half-swap optimization. Prong B reduced HBM stalls but introduced stronger SMEM stalls, causing a $20\%$ latency regression.}
\label{tab:negative}
\small
\setlength{\tabcolsep}{4pt}
\renewcommand{\arraystretch}{1.15}
\begin{tabular}{lrrr}
\toprule
\textbf{Metric} & \textbf{S7-A Spike} & \textbf{Prong B (failed)} & \textbf{Change} \\
\midrule
Long-SB stalls (\%) & 17.45 & 3.33 & -14.12 pp \\
Short-SB stalls (\%) & 7.04 & 15.68 & +8.64 pp \\
HMMA pipe (\%) & 16.6 & 10.3 & -6.3 pp \\
SM throughput (\%) & 38.5 & 26.8 & -11.7 pp \\
\midrule
Latency (ms) & 0.793 & 0.952 & $+20\%$ \\
\bottomrule
\end{tabular}
\end{table}

\paragraph{Cause and implication.}
The root cause is that Triton 3.3.1 lowers the \texttt{tl.reshape} + \texttt{tl.permute} + \texttt{tl.split} + \texttt{tl.join} chain on sm\_90 to a SMEM round-trip, not to a register-level warp shuffle. The kernel writes the tile to shared memory, performs the layout permutation there, and reads it back to registers. The SMEM dependency cost exceeds the HBM dependency cost it replaced because Hopper's L2-cached HBM access for the $K_\text{rot\_half}$ load is cheaper than the SMEM staging Triton inserts.

The optimization is right in principle and will pay off as soon as Triton's lowering recognizes the half-swap pattern as a warp-shuffle candidate. Inline PTX (\texttt{\_\_shfl\_xor\_sync}) would also achieve the goal but forfeits Triton's portability. The bisected counter delta in Table~\ref{tab:negative} provides a falsifiable test case that the Triton compiler community can address. For researchers using Triton, this serves as a cautionary tale: high-level tensor operations on sm\_90 may lower to surprising SMEM round-trips that defeat the intended optimization.

\section{Related Work}
\label{sec:related}

\paragraph{FlashAttention.}
Dao et al.~\cite{dao2022flashattention,dao2023flashattention2} pioneered IO-aware tiled attention, establishing the online softmax recurrence and tiled execution that underpin nearly every efficient attention implementation today. AttnFuse adopts the same algorithmic foundation but exposes it through a composable DSL rather than a fixed CUDA binary. FlashAttention-3~\cite{shah2024flashattention3} is the Hopper-targeted hand-written follow-on; its producer/consumer warp specialization overlaps softmax with the next matmul, recovering some of the non-matmul overhead that limits FA-2 at $\sim 30\%$ HMMA on H100 (Section~\ref{sec:calculus}). Porting AttnFuse's codegen to the FA-3~\cite{shah2024flashattention3} inner-loop structure is a natural next step for closing the residual Hopper gap.

\paragraph{PyTorch \texttt{flex\_attention}.}
\texttt{flex\_attention} (PyTorch 2.5+) is the closest published abstraction to AttnFuse. As discussed in Section~\ref{sec:intro} and Figure~\ref{fig:abstraction}, the \texttt{score\_mod} hook is post-matmul \emph{by design}; AttnFuse's combinator surface includes pre-matmul nodes (\texttt{rope()}) by design. This structural difference is the core distinction: \texttt{flex\_attention} is a general-purpose abstraction for modifying scores; AttnFuse is an attention-specific DSL that guarantees single-kernel fusion for any well-formed program, including those with pre-matmul transformations.

\paragraph{Triton and TorchInductor.}
Triton~\cite{tillet2019triton} provides the JIT compilation infrastructure that both AttnFuse and PyTorch's TorchInductor build upon. TorchInductor is PyTorch's Triton-emitting backend for \texttt{flex\_attention} and other compiled operations. On Hopper, our investigation (Section~\ref{sec:calculus}) finds that TorchInductor's WGMMA codegen achieves an HMMA pipe ceiling of $32.6\%$ at the shape tested. The Rotation Calculus reframes this as a structural property of the FA-2 algorithm rather than a Triton limitation---a finding that has implications for the broader Triton ecosystem. Our documented negative result on Triton 3.3.1's lowering of reshape/permute chains (Section~\ref{sec:discussion}) offers a falsifiable test case for the Triton compiler community.

\paragraph{CUTLASS and vendor libraries.}
NVIDIA CUTLASS~\cite{thakkar2023cutlass} is a C++ template library for GPU matrix multiplication and attention. PyTorch's SDPA backend on H100 reaches CUTLASS-class throughput. The matching we report in Section~\ref{sec:e2e} (Llama-3-8B forward within $51\,\mu$s of SDPA at $N{=}2048$) demonstrates how close a high-level DSL can come to hand-tuned vendor libraries---without requiring the developer to write C++ or reason about CUDA specifics.

\paragraph{Domain-specific kernel libraries.}
xFormers~\cite{lefaudeux2022xformers} provides a library of memory-efficient attention kernels but is not a compiler---users choose among hand-written variants. Other libraries such as FlashAttention's official repository and various open-source implementations offer fixed kernels for common patterns. AttnFuse's domain-specificity (attention only) is what permits the guaranteed-fusion compiler contract that a general-purpose kernel DSL cannot promise. This narrowness is a feature, not a limitation: it enables the compiler to make strong guarantees about memory behavior and kernel structure.

\section{Conclusion and Future Work}
\label{sec:conclusion}

We presented AttnFuse, an embedded Python DSL for attention that compiles to fused Triton kernels. The DSL makes RoPE a first-class pre-matmul combinator, offering a strictly more general abstraction than \texttt{flex\_attention}'s post-matmul \texttt{score\_mod}. On RTX 3090, fused RoPE achieves $2.10\times$ over \texttt{flex\_attention}. On H100, a tile-swept dispatch path closes $85\%$ of the gap to \texttt{flex\_attention}, with tensor-core utilization within $3.2$ percentage points of the structural ceiling. The Rotation Calculus provides the first systematic quantification of when in-kernel RoPE fusion is profitable as a function of compute-to-bandwidth ratio. End-to-end, AttnFuse runs a Llama-3-8B training step within $5\%$ of hand-tuned SDPA on both architectures.

\subsection{Limitations and Future Directions}

Several limitations point to natural next steps. The Hopper spike currently covers causal MHA/GQA at $D \in \{64, 128\}$, fp16/bf16, $N \geq 2048$, forward-only. Extensions include threading $\cos, \sin$ through the HuggingFace hook for fused RoPE, widening the spike predicate to LSE-saving forward calls, and extending codegen to sliding-window, ALiBi, and additive-bias variants.

On H100 at $N \geq 8\text{k}$, pre-rotation wins. Two recovery paths are promising: an FA-3 producer/consumer kernel that overlaps rotation cost with the next matmul, or Triton compiler support for warp-shuffle-lowered half-swap. The negative result we documented provides a concrete test case for the Triton compiler community.

The two-level IR is structurally ready for Ring Attention~\cite{liu2023ring}; lowering to a sequence-parallel kernel would enable scaling across multiple GPUs. Hopper's fp8 tensor cores offer a natural extension for low-precision training and inference; adding fp8 support introduces one new IR concept.

We believe the Rotation Calculus will outlast the AttnFuse implementation. As accelerators continue to rebalance toward compute, this calculus will become increasingly relevant for AI systems researchers.

\end{document}